%% file: xad_sm.tex
\documentclass{ifac/ifacconf}

\usepackage{graphicx}      
\usepackage{natbib}        
\usepackage{comment}
\usepackage{xcolor}
\usepackage{subcaption}
\usepackage{float}
\usepackage{url}
\usepackage{tabu} 
\usepackage{booktabs}

\begin{document}
\begin{frontmatter}

\title{Towards Explainable Anomaly Detection in Shared Mobility Systems\thanksref{footnoteinfo}} 

\thanks[footnoteinfo]{This work was partially carried out within the Italian National Center for Sustainable Mobility (MOST) and received funding from NextGenerationEU (Italian NRRP – CN00000023 - D.D. 1033 17/06/2022 - CUP C93C22002750006).}

\author{Elnur Isgandarov,}
\author{Matteo Cederle,} 
\author{Federico Chiariotti,} 
\author{Gian Antonio Susto}

\address{University of Padova, 
   Italy (e-mail: elnur.isgandarov@studenti.unipd.it, matteo.cederle@phd.unipd.it, federico.chiariotti@unipd.it, gianantonio.susto@unipd.it).}

\begin{abstract}
Shared mobility systems, such as bike-sharing networks, play a crucial role in urban transportation. Identifying anomalies in these systems is essential for optimizing operations, improving service reliability, and enhancing user experience. This paper presents an interpretable anomaly detection framework that integrates multi-source data, including bike-sharing trip records, weather conditions, and public transit availability. The Isolation Forest algorithm is employed for unsupervised anomaly detection, along with the Depth-based Isolation Forest Feature Importance (DIFFI) algorithm providing interpretability. Results show that station-level analysis offers a robust understanding of anomalies, highlighting the influence of external factors such as adverse weather and limited transit availability. Our findings contribute to improving decision-making in shared mobility operations.
\end{abstract}


\end{frontmatter}

\input{sections/intro}

\input{sections/related_works}

\input{sections/method}

\input{sections/experiments}

\input{sections/conclusions}

\bibliography{xad_sm}


\end{document}

%% file: sections/intro.tex
\section{Introduction}
\label{sec:intro}

Shared mobility systems, such as bike-sharing networks, have become integral to urban transportation, offering cost-effective and environmentally friendly alternatives to private vehicles and providing a solution to mass transit's last mile problem, as discussed by~\cite{yang2019spatiotemporal}. However, their optimization and management requires an understanding of complex demand patterns, which are influenced by a combination of interacting factors that~\cite{zhang2023user} studied, such as weather conditions, mass transit integration, the urban fabric of the neighborhood, and special events such as concerts, sporting events, and public holidays, as shown by~\cite{weinreich2023bike}. This complex tapestry can obfuscate anomalies: fluctuations in demand patterns or changes in the balance of traffic to and from some stations may lead to system inefficiencies and reduce service availability, which, as~\cite{chiariotti2018dynamic} show, is the target of bike-sharing system optimization. Understanding and detecting anomalies in shared mobility data is thus crucial for improving operational efficiency and maximizing service availability for users, and an effective anomaly detection (AD) system can
improve rebalancing and maintenance, reducing operational costs for the service.

Existing AD approaches in shared mobility typically focus on user behavior or system-wide trends. However, they often suffer from two key limitations: a lack of interpretability, making it difficult to understand why certain trips or stations are flagged as anomalous~(\cite{xu2023explainable}), and reliance on labeled data, which is generally unavailable in real-world deployments~(\cite{damicantonio2024utrans}).

To address these challenges, we propose an unsupervised anomaly detection framework that integrates multi-source data, including bike-sharing trip records, weather conditions, public transit availability, neighborhood data, and holiday schedules. We employ the Isolation Forest algorithm proposed by~\cite{liu2008isolation} to identify anomalous mobility patterns and leverage Depth-based Isolation Forest Feature Importance (DIFFI), proposed by \cite{carletti2023diffi}, to interpret the results. We consider a station-focused approach, analyzing anomalies for a given docked bike-sharing station and a given time, and highlight the potential benefits of this approach for Mobility-as-a-Service (MaaS) systems.
The key contributions of this work are as follows:
\begin{enumerate}
    \item We propose an unsupervised anomaly detection framework for shared mobility systems which integrates heterogeneous data sources, and we test it in a real-world dataset.
    \item We apply the DIFFI algorithm to interpret the results, providing insights into the factors originating anomalies in MaaS systems.
    \item We perform a neighborhood-level spatial analysis, showing how anomaly detection can serve as an early-warning system for potential mobility disruptions.
    \item We present a case study on anomaly spikes, showing how explainability methods can shed light on the effect of external factors such as weather and transit availability on mobility patterns.
\end{enumerate}

The remainder of this paper is organized as follows: Section \ref{sec:relworks} provides an overview of related works on MaaS systems and anomaly detection for shared mobility. Section \ref{sec:method} details our method, including data sources, feature engineering, and the proposed AD framework. Section \ref{sec:exp} presents and discusses the results, focusing on key findings and interpretability insights. Finally, Section \ref{sec:conc} concludes the paper and outlines future research directions.

%% file: sections/related_works.tex
\section{Related Work}
\label{sec:relworks}

The interpretability of unsupervised anomaly detection models is a crucial property to be able to understand and evaluate the results in a variety of domain: common lightweight techniques like Isolation Forest (IF) are scalable and efficient, but act as black boxes, limiting the resulting understanding of the causes of anomalous patterns. \cite{carletti2023diffi} proposed DIFFI as an explainability technique tailored for IF, which quantifies the effect of individual features on anomaly detection by computing local and global feature importance scores. 

In this work, we consider a shared mobility application, which has been extensively studied thanks to the availability of bike sharing and traffic data. However, urban mobility is a highly complex phenomenon, which is interlinked with all aspects of city life: firstly, shared mobility systems are not used in isolation, as \cite{schimohr2021spatial} show, but rather complement existing mass transit systems, taking on different roles in different cities. Secondly, the urban fabric and geographic features play a significant role: as~\cite{weinreich2023bike} show, European cities, which often have pedestrianized urban cores and mixed residential and commercial zoning in the suburbs, have very different usage patterns with respect to US cities, which usually have a downtown business district surrounded by purely residential suburbs. The presence of hills and rivers also represents an important barrier that shapes mobility around it. Moreover, weather conditions strongly influence ridership, and the intensity and direction of these effects can differ across individual stations, necessitating localized analysis as argued by~\cite{kim2018investigation}. The temporal dimension is also important, as \cite{etienne2014model} and several others have argued: rush hour peaks are often clearly visible in usage data, and stations have different balances between incoming and outgoing traffic at different times, even without accounting for weekly and seasonal cycles.

In this complex scenario, detecting anomalies is a daunting task, which has only been undertaken in the past few years. A first attempt by~\cite{el_sibai2018spatial} only considered the spatial dimension, trying to find stations that had significant differences in their usage patterns with their immediate neighbors. This approach complements the clustering method used by~\cite{etienne2014model} and~\cite{weinreich2023bike} by considering local outliers. Other works have taken a different approach, concentrating on the temporal dimension:~\cite{lam2019detecting} developed a method to detect external events by identifying and analyzing significant deviations in ridership patterns, but the authors rarely managed to link anomalies to specific events, struggling to obtain clear-cut explanations of their results. Finally,~\cite{liu2022data} proposed a two-stage functional outlier detection method for identifying abnormal patterns in real-time station occupancy data. This work showed good performance at identifying and predicting imbalances, with clear benefits for system operators, but did not attempt to find explanations for the identified anomalies.

Despite the promise of these approaches, the gap between the models, theories, and explanations identified in the urban transport literature and quantifiable anomalies in the shared mobility data persists. Due to the size of the datasets, which limit the use of computationally intense approaches, and the complexity of the environment, which requires the combination of different spatio-temporal datasets and prevents easy labeling, the proper identification and explanation of anomalies is difficult, and distinguishing true anomalies from short-term fluctuations is challenging. Our work represents a first step toward the explainability of these anomalies, with some encouraging results on a thorny and still mostly unexplored topic.

%% file: sections/method.tex
\section{Methodology}
\label{sec:method}

This section outlines the methodology used to detect anomalies in shared mobility data. The proposed framework integrates multi-source data, applies feature engineering to extract meaningful variables, and employs an interpretable unsupervised anomaly detection model to extract meaningful anomalies.

This study focuses primarily on the BlueBikes system, a dock-based sharing system that has been active in Boston, MA (USA) since the summer of 2011. BlueBikes regularly publishes trip records that include the start and end stations and timestamps for all recorded trips. The data is available for the past $10$ years, and includes basic demographic information about the user taking the trip. We selected the month of January 2023 for our analysis, as all sources provided the required data for that period.

However, the need to consider the most important factors in the literature led us to include different data sources to expand the context of our AD method:
\begin{itemize}
    \item Hourly meteorological conditions, such as temperature, precipitation, and wind speed, from the open weather and climate data aggregator Meteostat;
    \item Mass transit data from the Massachusetts Bay Transportation Authority (MBTA), which publishes transit schedules and stop locations for the whole area;
    \item The geographic boundaries of census blocks and neighborhoods in Boston and Cambridge and the official public holiday calendar.
\end{itemize}

These datasets include the most common spatial factors influencing bike sharing traffic, i.e., the interaction with mass transit and the urban fabric, and the most common temporal ones, i.e., holidays and the weather, which affect both mobility patterns and the willingness of users to cycle instead of choosing another transport mode.

\begin{table*}[htb]
\caption{Features used as inputs for the AD algorithm.}
\centering
\scriptsize{
\renewcommand*{\arraystretch}{1.05}
\begin{tabular}{c|ll}
\toprule
Data type & Source & Features\\
\midrule
Bike sharing data & \url{https://bluebikes.com/system-data} & Traffic load and type, avg. distance, avg. duration, avg. speed, user type\\
Timing & & Hour, day of the month, weekday\\
Weather & \url{https://meteostat.net/} & Avg. temperature, avg. precipitation, avg. wind speed, \\ & & ``coco'' (categorical representation of weather conditions)\\
Mass transit & \url{https://mbta-massdot.opendata.arcgis.com/} & Nearby transit stop number\\
Spatial & \url{https://data.boston.gov/} & Neighborhood type\\
                    \bottomrule
\end{tabular}\label{tab:features}}
\end{table*}

\begin{figure*}[t]
    \centering
    \begin{subfigure}[b]{0.48\textwidth}
        \centering
        \includegraphics[width=\textwidth]{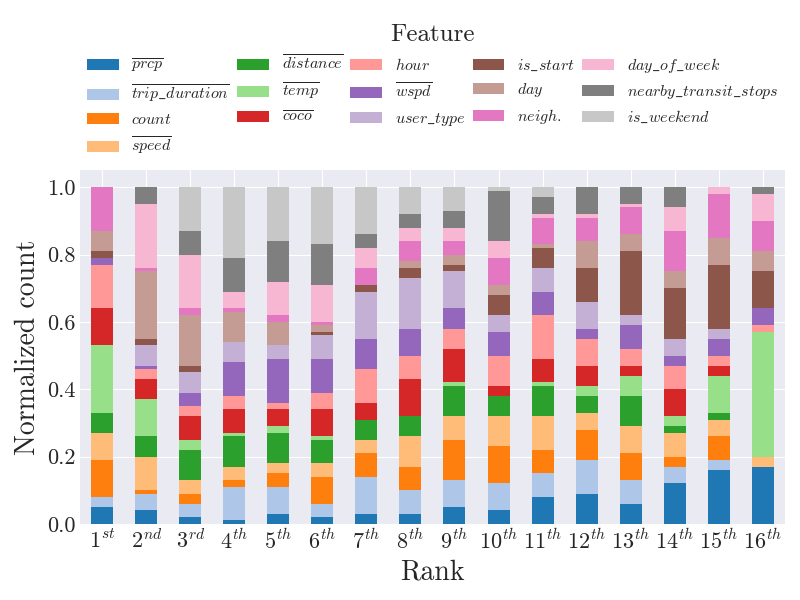}
        \caption{Local-DIFFI}
        \label{fig:ldiffi}
    \end{subfigure}
    \hfill
    \begin{subfigure}[b]{0.48\textwidth}
        \centering
        \includegraphics[width=\textwidth]{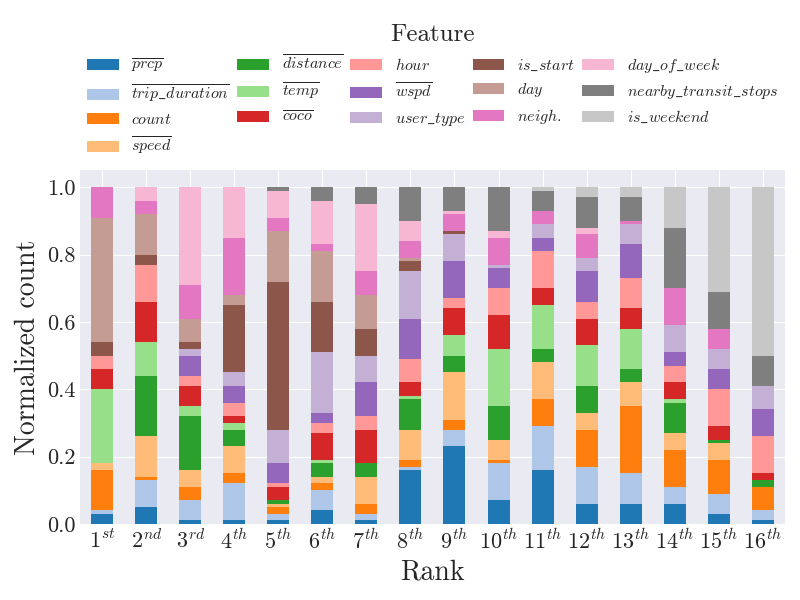}
        \caption{SHAP}
        \label{fig:shap}
    \end{subfigure}
    
    \caption{Feature rankings based on Local-DIFFI and SHAP scores.}
    \label{fig:comparison}
\end{figure*}

The data included in the AD input are aggregated by station, as individual trips exhibit a large variability and are much harder to interpret. For a specific hour of a specific day, we consider the incoming or outgoing traffic at a station, and extend the dataset by considering the ratio between subscribers and on-demand users, as well as the average distance, duration, and linear speed of the trips. We also include information about the timing of the considered hour, in order to account for circadian, weekly, and seasonal cycles. Weather data is also included, as well as basic spatial information such as the neighborhood type and the number of transit stops within a $300$~m radius. The features and data sources are summarized in Table~\ref{tab:features}.

The anomaly detection framework we used is based on the IF technique, which recursively partitions the datasets to isolate abnormal instances. IF is designed to handle high-dimensional unlabeled datasets by building random binary trees and evaluating how often each point is isolated by computing an anomaly score. Furthermore, we exploited the DIFFI algorithm to explain the anomaly scores, computing local and global feature rankings: in this work, we used local feature importance scores to evaluate individual predictions in an efficient manner, comparing it with the outputs of the SHAP algorithm. The main advantage of DIFFI is its low computational complexity: the average computational time per sample\footnote{Experiments were run on a standard consumer laptop equipped with an 2,6 GHz 6-Core Intel Core i7 CPU and 16 GB RAM.}) for DIFFI was $0.293$~s, in contrast to the $5.955$~s per sample required by SHAP.

The model was initially applied to trip-level data to identify individual outliers based on duration and distance and spot atypical trips, as well as tuning the contamination parameter, determining the fraction of detected anomalies. However, individual trip data is subject to a higher randomness and limited interpretability, as it is hard to ascertain whether anomalous trips were due to the presence of atypical users or to larger-scale patterns. To address these challenges, we used the individual trip-level parameter tuning to examine anomalies over the aggregated traffic for a certain period of time at a given station, which is much less sensitive to the behavior of individual users. The full interpretation and analysis was then performed on these aggregated station-level data to identify anomalous traffic patterns in time and space.
This two-stage approach ensures both effective anomaly identification and meaningful interpretation of the results.

%% file: sections/experiments.tex
\section{Experimental Results}
\label{sec:exp}

This section presents the findings from our explainable anomaly detection approach applied to shared mobility. At the beginning, the focus is put on evaluating the significance of various features in generating anomalies. Subsequently, a spatial analysis will be conducted both for neighborhoods and individual stations to uncover the spatial dynamics associated with such anomalies. The section will then conclude with a case study examining specific peaks in the frequency of anomalies, offering an insightful example of how the DIFFI algorithm can be leveraged for the interpretation of anomalies. Furthermore, this analysis can also be used to provide actionable insights for service providers\footnote{Code available at \url{https://github.com/elnurisg/anomaly_detection_for_shared_mobility}.}.

\subsection{Feature Importance Analysis}
As a first analysis, we applied the Local-DIFFI algorithm to provide an interpretation of individual predictions. As we can see from Figure~\ref{fig:ldiffi}, our investigation revealed that features such as the temperature, neighborhood, and day of the week consistently appear among the top ranked features useful for predicting anomalies, confirming the substantial evidence that these cyclical patterns tend to be the most important factors determining urban mobility.
Additionally, transit accessibility emerged as a critical factor; stations with limited nearby transit options exhibited higher rates of anomalies. This result can be attributed to the importance of multi-modal trips in bike sharing traffic patterns, as most commuters tend to use shared micromobility as a last mile solution in conjunction with some form of mass transit. Additionally, lower-occupancy or underprivileged neighborhoods, which have a lower density of mass transit stops, may have different usage patterns linked to socio-economic factors.

As we discussed in Section \ref{sec:relworks}, the majority of real-world bike sharing datasets do not label anomalous records such as, e.g., rebalancing operations or mechanical failures in the bikes, making the validation of AD algorithms extremely challenging. However, the comparison between Local-DIFFI and SHAP~(\cite{lundberg2017unified}), visible in Figure \ref{fig:comparison}, provides some additional robustness to the results: both DIFFI and SHAP identified similar features as the most important ones, indicating that the results are rather stable with respect to the choice of the specific explainability framework.

\begin{figure}[t]
\begin{center}
\includegraphics[width=8.8cm]{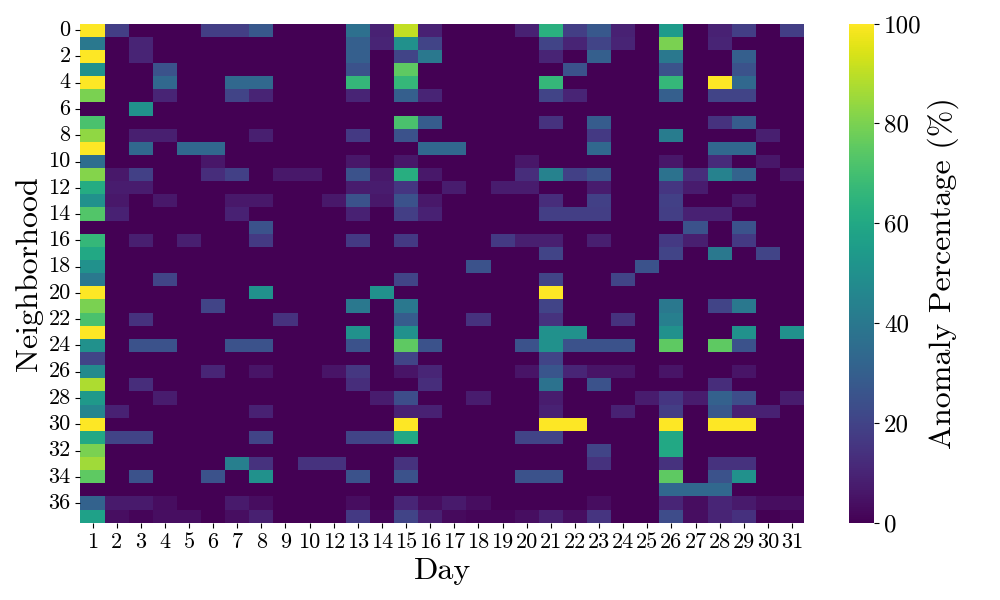}
\caption{Percentage of anomalous stations by neighborhood and day of the month (January 2023).} 
\label{fig:neigh1}
\end{center}
\end{figure}

\begin{figure}[t]
\begin{center}
\includegraphics[width=8.8cm]{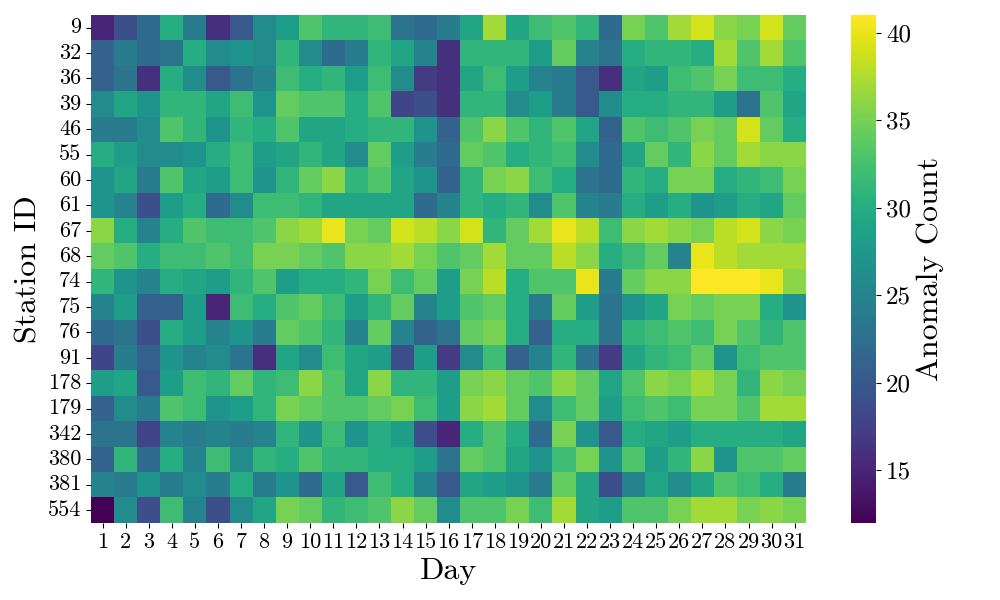}
\caption{Anomaly counts for the first 20 stations and day of the month (January 2023).} 
\label{fig:neigh2}
\end{center}
\end{figure}

\begin{figure}[t]
\begin{center}
\includegraphics[width=8.4cm]{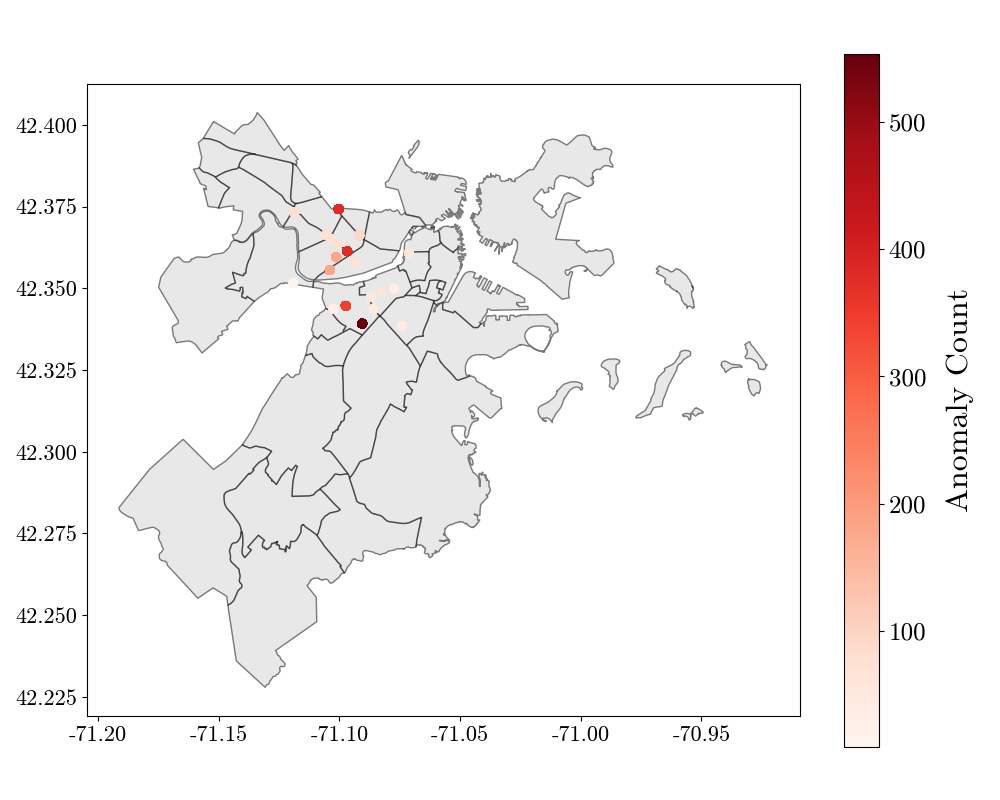}
\caption{Spatial distribution of the 20 stations with highest anomaly count in the city of Boston, MA.} 
\label{fig:neigh3}
\end{center}
\end{figure}
\subsection{Neighborhood and Station-Level Spatial Analysis}
To further understand the spatial dynamics of anomalies in shared mobility systems, a neighborhood and station-level analysis was conducted to uncover localized disruptions and identify high-risk zones. 
To offer a normalized comparison across areas, we reported the percentage of anomalous stations for each neighborhood and day  in Figure \ref{fig:neigh1}. As expected, New Year's Day exhibited widespread anomalous activity due to large-scale events, underscoring external influences on anomaly rates.

Interestingly, Martin Luther King Jr. Day, a federal holiday which occurred on January 16th, was not associated with large-scale anomalies. However, the preceding day, Sunday, January 15th, had the highest wind speed throughout the dataset. Thursday, January 26th also had a high frequency of anomalous stations, and it was the only day of the month with a heavy rainfall ($2.6$~cm throughout the day, almost double the amount for any other day). Finally, neighborhood 30, i.e., \textit{Strawberry Hill}, experienced the highest number of anomaly peaks throughout the month. Interestingly, nearly all of these occurred during the weekends, likely due to the limited availability of public transportation in the area on those days—a known issue in this particular neighborhood.

\begin{figure*}[t]
    \centering
    \begin{subfigure}[b]{0.48\textwidth}
        \centering
        \includegraphics[width=\textwidth]{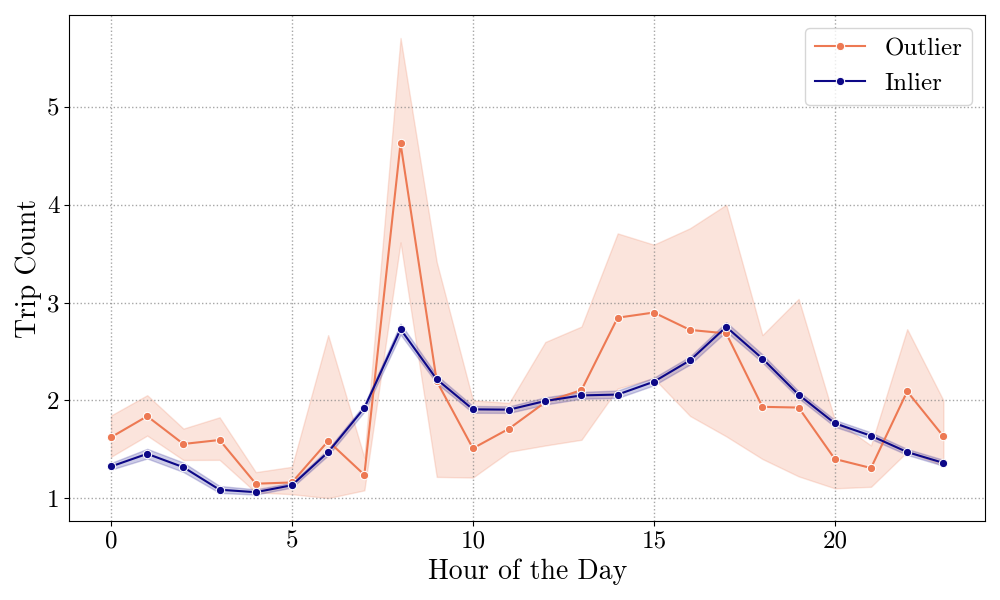}
        \caption{Trips count over hours of the day}
        \label{fig:hourly}
    \end{subfigure}
    \hfill
    \begin{subfigure}[b]{0.48\textwidth}
        \centering
        \includegraphics[width=\textwidth]{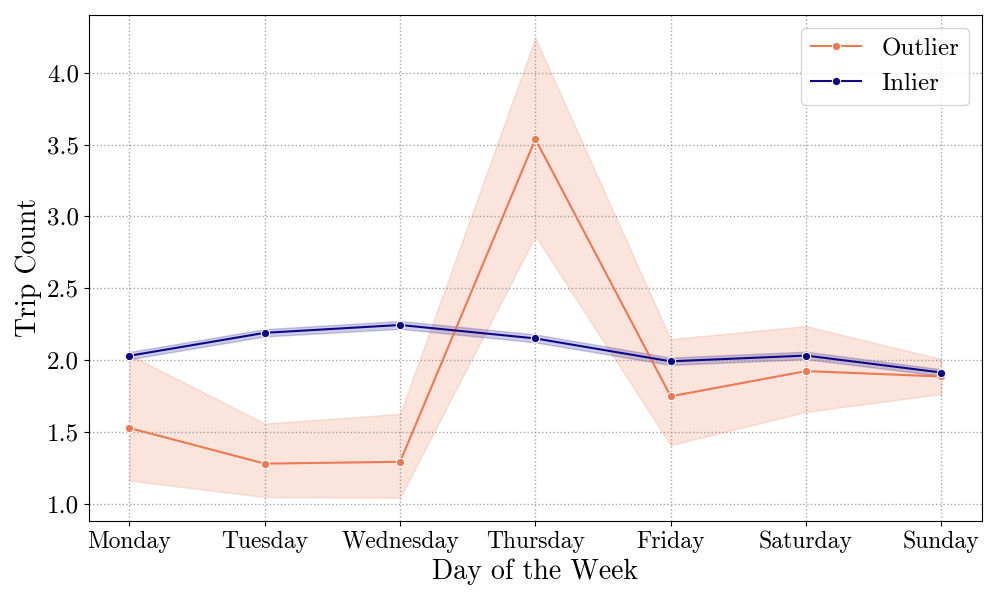}
        \caption{Trips count over days of the week}
        \label{fig:daily}
    \end{subfigure}
    
    \caption{Temporal analysis of anomalies: hourly (a) and daily (b) trip distributions. The shaded areas represent the 95$\%$ confidence interval of the predictions.}
    \label{fig:timean}
\end{figure*}

\begin{figure}[h]
\begin{center}
\includegraphics[width=8.8cm]{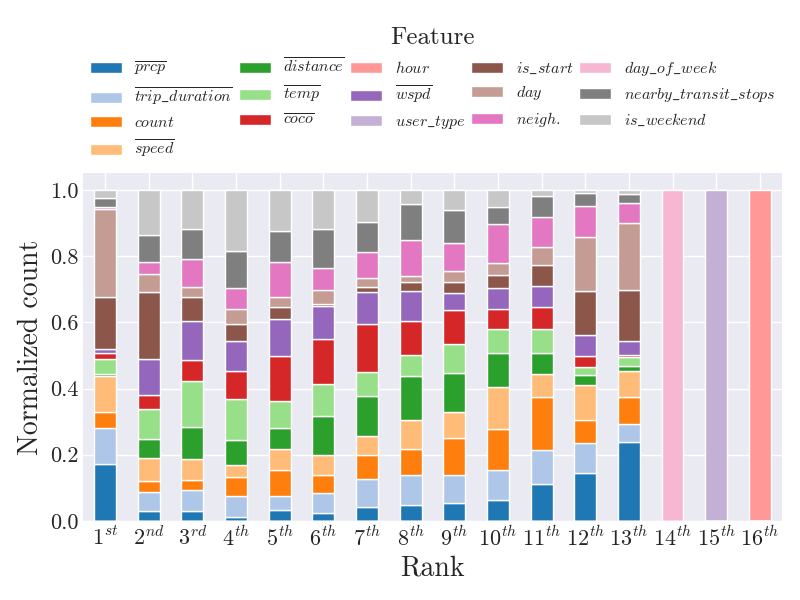}
\caption{Feature rankings based on Local-DIFFI scores for anomalies on 8 a.m. on Thursdays.} 
\label{fig:ldiffi8am}
\end{center}
\end{figure}

At the station level, a focused examination of the first $20$ stations, ranked by anomaly count, provided more granular insights. As we can see from Figure~\ref{fig:neigh2}, the heatmap highlights specific stations with recurrent anomalies, indicating potential localized factors or recurring events driving these patterns. To better understand the spatial context of these high anomaly stations, Figure~\ref{fig:neigh3} maps their locations across the city. We note that stations with frequent anomalies are clustered in an area, with the most frequent anomalous traffic occurring at stations near the Longwood medical campus, which can explain frequent anomalies, as patients, doctors, and medical students and interns represent a large population that does not follow standard commuting hours. The other stations with frequent anomalies are clustered near the Harvard and MIT campuses in Cambridge, north of the Charles River.

\subsection{Case Study: Anomaly Spikes and External Influences}

Finally, to examine how anomalies are distributed across different time periods, we performed a temporal analysis, as shown in Figure \ref{fig:timean}. A notable spike in anomalies occurs around $8$~a.m., aligning with the morning commute period, while Thursdays also stand out as the weekday with the most anomalies. These observations indicate that unusual activity tends to cluster around specific temporal patterns, suggesting potential external influences that merit further investigation. After a comparative analysis, omitted here for the sake of brevity, we observed that the majority of $8$~a.m. anomalies occurred on Thursdays, and conversely Thursday anomalies often occurred at $8$~a.m. To provide interpretations and try to explain the origin of such anomalies, we applied the Local-DIFFI algorithm only for the anomalies at 8 a.m. on Thursdays. The results, shown in Figure \ref{fig:ldiffi8am}, indicate that public transit availability and environmental conditions were the key contributors to the anomalies: wind speed and precipitation increase their importance, and trip duration, speed, and the presence of nearby transit stops do so much more.

Concluding this section, we remark that these findings demonstrate how AD can serve as an early-warning system for potential mobility disruptions, providing actionable insights for urban planners and transit operators.

%% file: sections/conclusions.tex
\section{Conclusion and Future Directions}
\label{sec:conc}

This study presented an explainable anomaly detection framework for shared mobility systems, focusing on bike-sharing networks. By integrating multi-source data, including bike-sharing trip records, weather conditions, public transit availability, neighborhood characteristics, and holiday schedules, the proposed approach enhances anomaly detection accuracy while maintaining interpretability. The application of IF, combined with DIFFI for feature importance analysis, provided meaningful insights into the underlying factors influencing anomalies, which can be exploited by shared mobility operators and urban planners. By identifying atypical patterns in bike usage, system operators can proactively address operational inefficiencies, optimize resource allocation, and improve user experience. Furthermore, integrating external factors in the analysis allowed for a more comprehensive understanding of mobility patterns, aiding strategic planning and infrastructure development.

Future research directions include the extension of the analysis to longer time periods and multiple cities, as well as the integration of additional external data sources into the model, such as demographic trends and real-time traffic patterns.